\documentclass[letterpaper]{article} 
\usepackage{aaai2026}  
\usepackage{times}  
\usepackage{helvet}  
\usepackage{courier}  
\usepackage[hyphens]{url}  
\usepackage{graphicx} 
\usepackage{natbib}  
\usepackage{caption} 
\usepackage{algorithm}
\usepackage{algorithmicx}
\usepackage{algpseudocode}
\usepackage{booktabs}
\usepackage{multirow}
\usepackage{amsmath}
\usepackage{amssymb}
\usepackage{longtable}
\usepackage{dashrule}
\usepackage{eso-pic}  

\newcommand{\think}[1]{\textcolor{blue}{\sffamily #1}}
\newcommand{\search}[1]{\textcolor{cyan}{\sffamily #1}}
\newcommand{\information}[1]{\textcolor{brown}{\sffamily #1}}
\newcommand{\answer}[1]{\textcolor{magenta}{\sffamily #1}}
\newcommand{\question}[1]{\textcolor{red}{\sffamily #1}}

\usepackage{newfloat}
\usepackage{listings}
\DeclareCaptionStyle{ruled}{labelfont=normalfont,labelsep=colon,strut=off} 
\floatstyle{ruled}
\newfloat{listing}{tb}{lst}{}
\floatname{listing}{Listing}
\usepackage{eso-pic}
\AddToShipoutPictureBG*{%
  \AtPageUpperLeft{%
    \put(60,-70){
      \noindent \includegraphics[height=1.0cm]{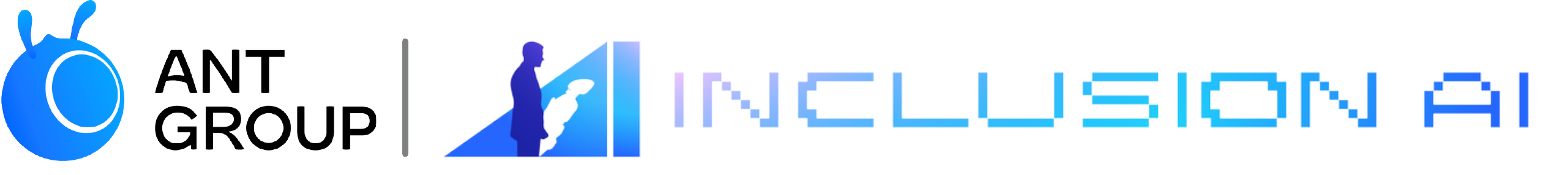}
    }%
  }%
}

\title{RAG-R1: Incentivizing the Search and Reasoning Capabilities of LLMs through Multi-query Parallelism}
\author{
    Zhiwen Tan\textsuperscript{\rm 1}, 
    Jiaming Huang\textsuperscript{\rm 1,2},
    Qintong Wu\textsuperscript{\rm 1},
    Hongxuan Zhang\textsuperscript{\rm 1,3},
    Chenyi Zhuang\textsuperscript{\rm 1}\thanks{Corresponding author.},
    Jinjie Gu\textsuperscript{\rm 1}
}
\affiliations{
    \textsuperscript{\rm 1}Inclusion AI, Ant Group\\
    \textsuperscript{\rm 2}Zhejiang University\\
    \textsuperscript{\rm 3}Nanjing University\\
    ender.tzw@antgroup.com, chenyi.zcy@antgroup.com

}

\usepackage{bibentry}

\begin{document}

\maketitle

\begin{abstract}
Large Language Models (LLMs), despite their remarkable capabilities, are prone to generating hallucinated or outdated content due to their static internal knowledge. 
While Retrieval-Augmented Generation (RAG) integrated with Reinforcement Learning (RL) offers a solution, these methods are fundamentally constrained by a single-query mode, leading to prohibitive latency and inherent brittleness.
To overcome these limitations, we introduce RAG-R1, a novel two-stage training framework centered around multi-query parallelism.
Our framework enables LLMs to adaptively leverage internal and external knowledge during the reasoning process while transitioning from the single-query mode to multi-query parallelism.
This architectural shift bolsters reasoning robustness while significantly reducing inference latency.
Extensive experiments on seven question-answering benchmarks confirm the superiority of our method, which outperforms the strongest baseline by up to 13.7\% and decreases inference time by 11.1\%.
\end{abstract}

\begin{links}
    \link{Code}{https://github.com/inclusionAI/AWorld-RL/tree/main/RAG-R1}
\end{links}

\section{Introduction}
Large Language Models (LLMs)~\citep{Survey_of_LLMs,deepseek-r1} have demonstrated remarkable capabilities in various domains, including mathematical reasoning, question answering, and code generation.
However, the knowledge encoded in these models is static, limiting their adaptability.
As a result, LLMs are susceptible to producing hallucinated or outdated responses~\citep{Survey_of_Hallucination,RAG_Reduce_Hallucination,self-rag} when dealing with complex or real-time issues, compromising their reliability. 
Therefore, it is essential to equip LLMs with access to external knowledge to ensure more accurate and grounded responses.

Retrieval-Augmented Generation (RAG)~\citep{RAG_A_Survey,RAG_Meeting_LLMs} is a widely adopted approach to address this issue, broadening the model's capability boundaries by integrating external knowledge into the generation process.
Early efforts~\citep{ITER-RETGEN,In_Context_RAG} in this area concentrated on prompt-based strategies that guide LLMs through question decomposition, query rewriting and multi-turn retrieval.
While effective, these approaches are constrained by the limitations inherent in prompt engineering.
Recent advances~\citep{IRCoT,Search-o1,self-rag,DeepRAG,CoRAG} have increasingly emphasized the integration of reasoning capabilities within RAG.
These approaches combine RAG with Chain of Thought (CoT) for step-by-step retrieval, or automatically generate intermediate retrieval chains that incorporate reasoning and employ Supervised Fine-Tuning (SFT) to enable learning of retrieval and reasoning.
However, new insights~\citep{SFT_Memorizes} indicate that these methods may cause models to memorize solution paths, thus constraining their generalization to novel scenarios.

\begin{figure}[t]
    \begin{minipage}[t]{0.5\linewidth}
        \centering
        \includegraphics[width=\textwidth]{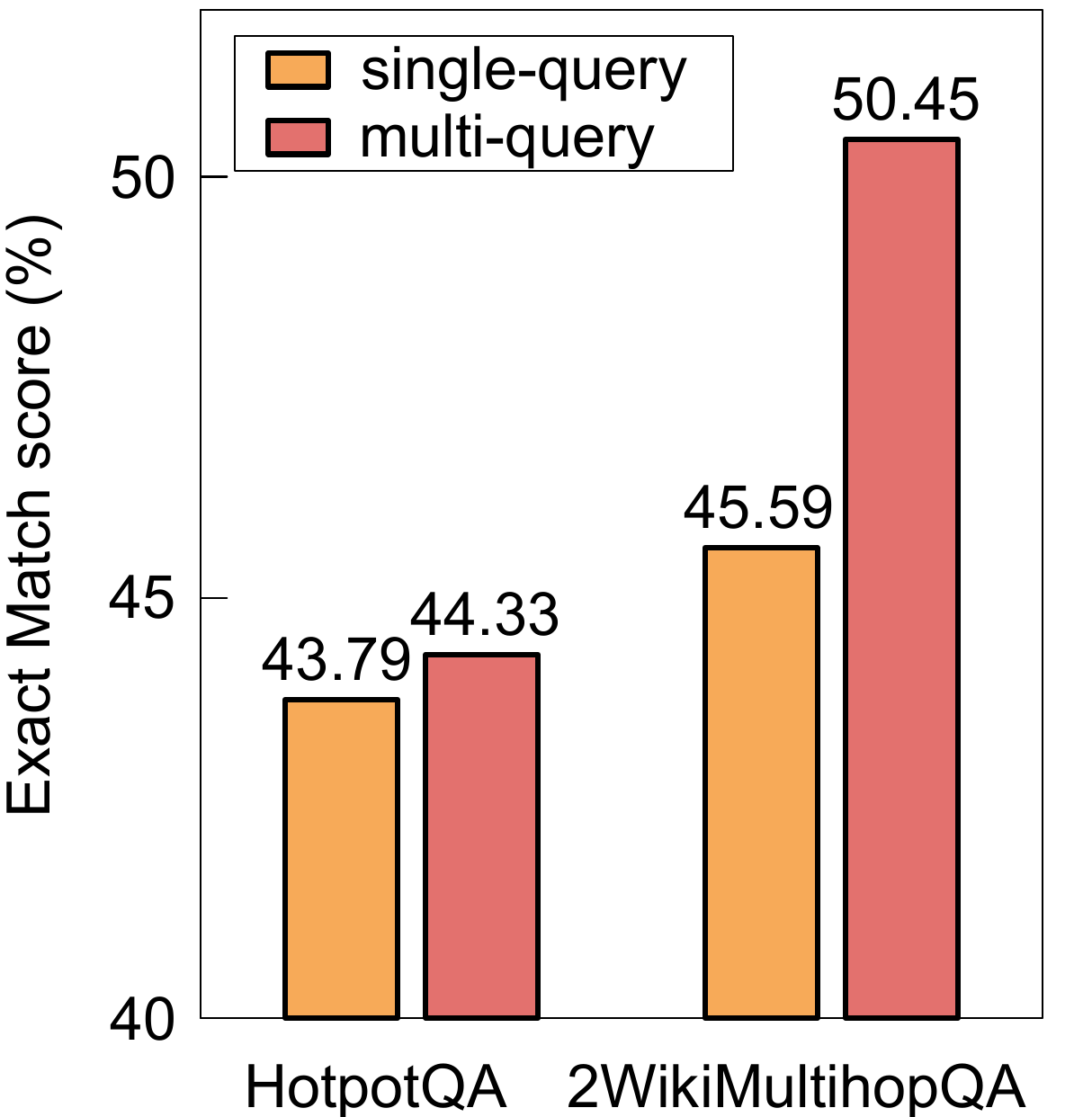}
        \centerline{(a) Exact Match score}
    \end{minipage}%
    \hfill
    \begin{minipage}[t]{0.5\linewidth}
        \centering
        \includegraphics[width=\textwidth]{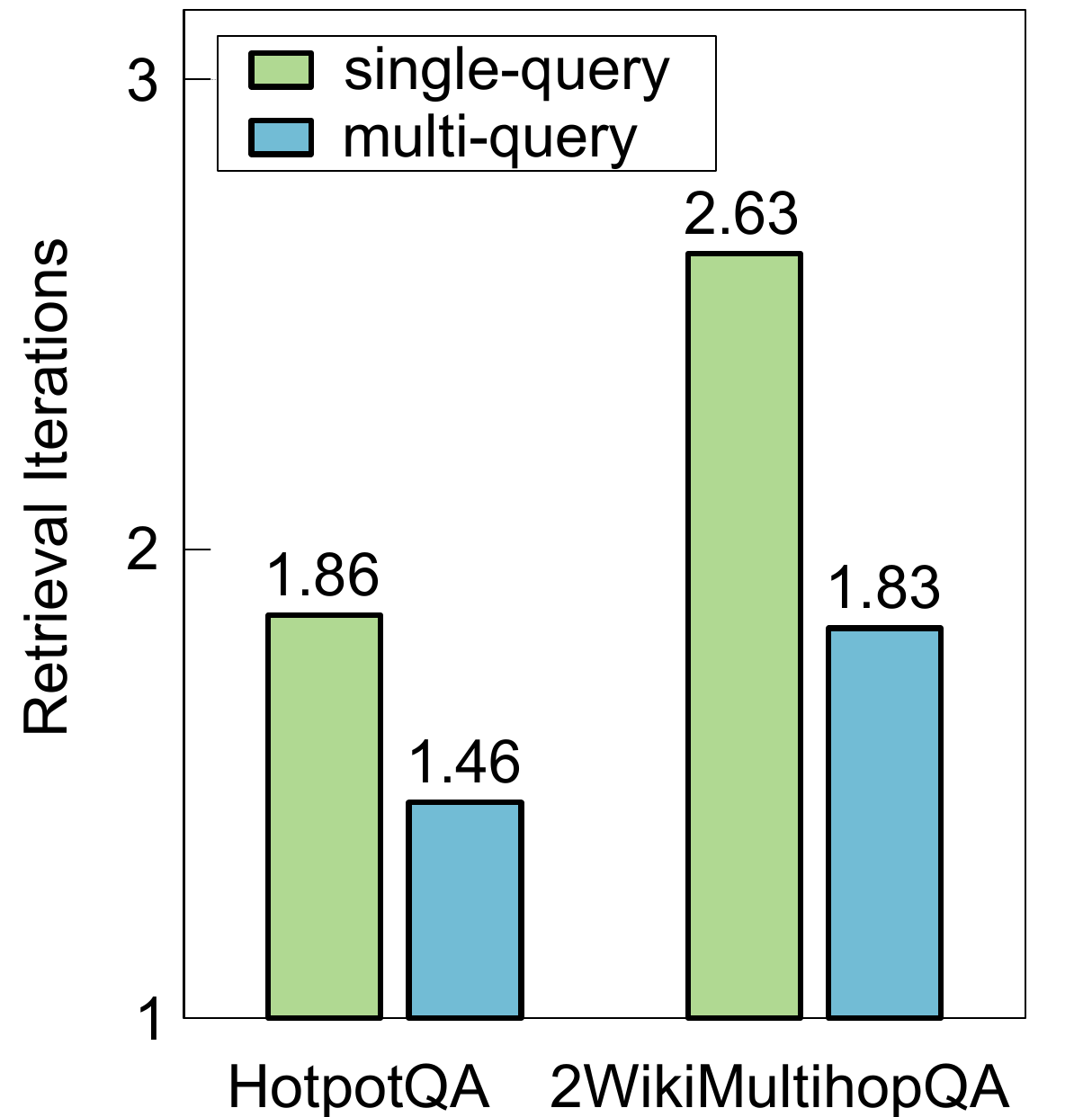}
        \centerline{(b) Retrieval Iterations}
    \end{minipage}
    \caption{Comparison of single-query and multi-query methods on Multi-Hop QA benchmarks based on Qwen2.5-72B-Instruct: (a) Model performance evaluated by Exact Match metric; (b) Average retrieval iterations during inference. The multi-query approach achieves a higher Exact Match score with fewer retrieval iterations.}
    \label{motivation}
\end{figure}

Recently, reinforcement learning (RL)~\citep{PPO} has demonstrated great potential in improving LLM performance by enhancing the reasoning capability.
Reasoning models such as OpenAI-o1~\citep{openai-o1} and Deepseek-R1~\citep{deepseek-r1} indicate that utilizing RL with outcome-based rewards can enhance the model's performance in mathematical and logical reasoning tasks. 
Within this paradigm, several works have explored enhancing the model's search ability during reasoning through RL. 
R1-Searcher~\citep{R1-Searcher} proposes a novel two-stage outcome-based RL approach designed to enhance the search capability of LLMs. 
Search-R1~\citep{Search-R1} introduces a novel RL framework that enables LLMs to interact with search engines in an interleaved manner with their own reasoning. 
Despite significant improvements, these RL-based methods struggle with stable training due to the restricted capabilities of cold-start models~\citep{deepseek-r1}.
Furthermore, existing methods generate only a single search query whenever external retrieval is required, which presents two significant challenges: 
(1) \textbf{Prohibitive Latency from Serial Execution.} The single-query architecture mandates a serial execution model for multi-hop reasoning, where each step must await the completion of the preceding retrieval-inference cycle. Consequently, latency accumulates with each turn, rendering the model impractical for latency-sensitive, real-world applications.
(2) \textbf{Inherent Brittleness and Knowledge Confinement.} The single-query approach is inherently brittle, as its entire reasoning trajectory depends on the success of each sequential step. An early, suboptimal query can lock the model into an unrecoverable path—a critical failure mode. This single-threaded process confines the model to a narrow slice of external knowledge, leaving it unable to navigate user ambiguity, explore alternative reasoning paths, or recover from an initial misstep. This fragility ultimately compromises its reasoning robustness and final performance.
We conducted a straightforward experiment based on Qwen2.5-72B-Instruct~\citep{qwen2.5} following \citet{Search-R1} to validate the aforementioned challenges.
As illustrated in Figure \ref{motivation}, we evaluated the model's performance and average retrieval count on two Multi-Hop Question Answering benchmarks: HotpotQA~\citep{yang2018hotpotqa} and 2WikiMultiHopQA~\citep{xanh2020_2wikimultihop}, comparing the generation of a single query to multiple queries in scenarios necessitating retrieval.
The findings indicate that the multi-query method enhances the model's performance and decreases retrieval iterations compared to the single-query approach, highlighting the limitations inherent in the single-query mode.

To address the aforementioned challenges, we propose RAG-R1, a novel training framework that enables LLMs to adaptively leverage internal and external knowledge during the reasoning process and enhances their reasoning capabilities.
We further expand the single-query mode to multi-query parallelism, an architectural shift that directly addresses prohibitive latency and inherent brittleness, thereby bolstering reasoning robustness and reducing inference time.
Specifically, the training framework contains two stages, i.e., Format Learning Supervised Fine-Tuning and Retrieval-Augmented Reinforcement Learning. 
In the first stage, we thoughtfully generate samples that integrate reasoning and search to equip LLMs with the ability to adaptively leverage internal and external knowledge during the reasoning process and respond in a think-then-search format. 
In the second stage, we employ outcome-based RL with a retrieval environment to enhance the model’s ability to reason effectively and dynamically access external knowledge for accurate question answering.
Building upon the training framework, we transition from the single-query mode to multi-query parallelism. By minimizing serial retrieval iterations, this approach drastically reduces overall inference time. Simultaneously, it arms the model with comprehensive evidence from multiple perspectives, directly bolstering its robustness against reasoning failures.

We conduct extensive experiments based on Qwen2.5-7B-Instruct~\citep{qwen2.5} to verify the effectiveness of our proposed method.
The results demonstrate its effectiveness, which achieves state-of-the-art performance on seven question-answering benchmarks.
In particular, our method utilizing multiple-query parallelism outperforms the strongest baseline by up to 13.7\% and decreases inference time by 11.1\%.
The contributions of this work are summarized as follows:
\begin{itemize}
  \item We propose RAG-R1, a novel training framework that empowers LLMs to adaptively leverage internal and external knowledge during the reasoning process and enhances their reasoning capabilities.
  \item We transition from the single-query mode to multi-query parallelism to directly address prohibitive latency and inherent brittleness. This architectural shift bolsters reasoning robustness while significantly reducing inference time.
  \item Extensive experiments demonstrate the effectiveness of our proposed method, which achieves state-of-the-art performance on seven question-answering benchmarks and significantly decreases retrieval counts and inference time.
\end{itemize}

\section{Related Work}
\label{related_work}
\textbf{Retrieval-Augmented Generation}
RAG enhances LLMs outputs by integrating external information with internal knowledge. 
Initial RAG methods employed static pipelines, coupling retrieved documents with LLMs via prompt engineering~\citep{replug,ralm, query-decomposing, multi-aspect-query} to control generation. 
While effective for information-sourcing tasks, these approaches struggle with complex, multi-faceted queries requiring iterative retrieval. 
Subsequent research developed dynamic RAG frameworks~\citep{adaptive-rag} that utilize LLM feedback for adaptive retrieval and generation control. 
Complementary efforts have focused on improving knowledge representation of retrieved documents~\citep{graph-rag, light-rag} from augmentation perspectives. 
Most recently, advances~\citep{Search-R1, Search-o1, R1-Searcher, zerosearch, masksearch} demonstrate significant scalability potential by training interaction trajectories to optimize LLM's use of search engines for complex problem-solving queries. 
Following these advances, our method further enhances the interaction between LLMs and retrievers by leveraging LLM's reasoning abilities.

\textbf{Learning to Search}
Recent advances in Learning to Search have evolved from static query and generation template via SFT to dynamic and reward-driven search using RL. 
SFT-based approaches have demonstrated success in improving LLM capabilities in instruction following~\citep{vif-rag, self-play, retrollm}, robustness~\citep{dpa-rag}, and domain-specific adaptation~\citep{raft}. 
However, these methods are limited by a tendency to memorize solution paths, which constrains generalization and scalability. 
RL-based methods~\citep{openai-o1,deepseek-r1,grpo,qwen3} offer a promising alternative by enabling autonomous reasoning and decision-making. 
Recent works~\citep{Search-o1, R1-Searcher, Search-R1} further integrate LLMs into real-time search workflows, allowing interaction with live search engines. 
Although these methods support autonomous search engine invocation, the efficacy of search-as-a-tool remains underexplored. 
To address this, our approach introduces a two-stage training framework that enables parallel query generation, with the aim of retrieving more comprehensive and diverse document contexts.

\section{Training Framework}
\label{training_framework}
\begin{figure*}[htbp] 
  \centering
  \includegraphics[width=\linewidth]{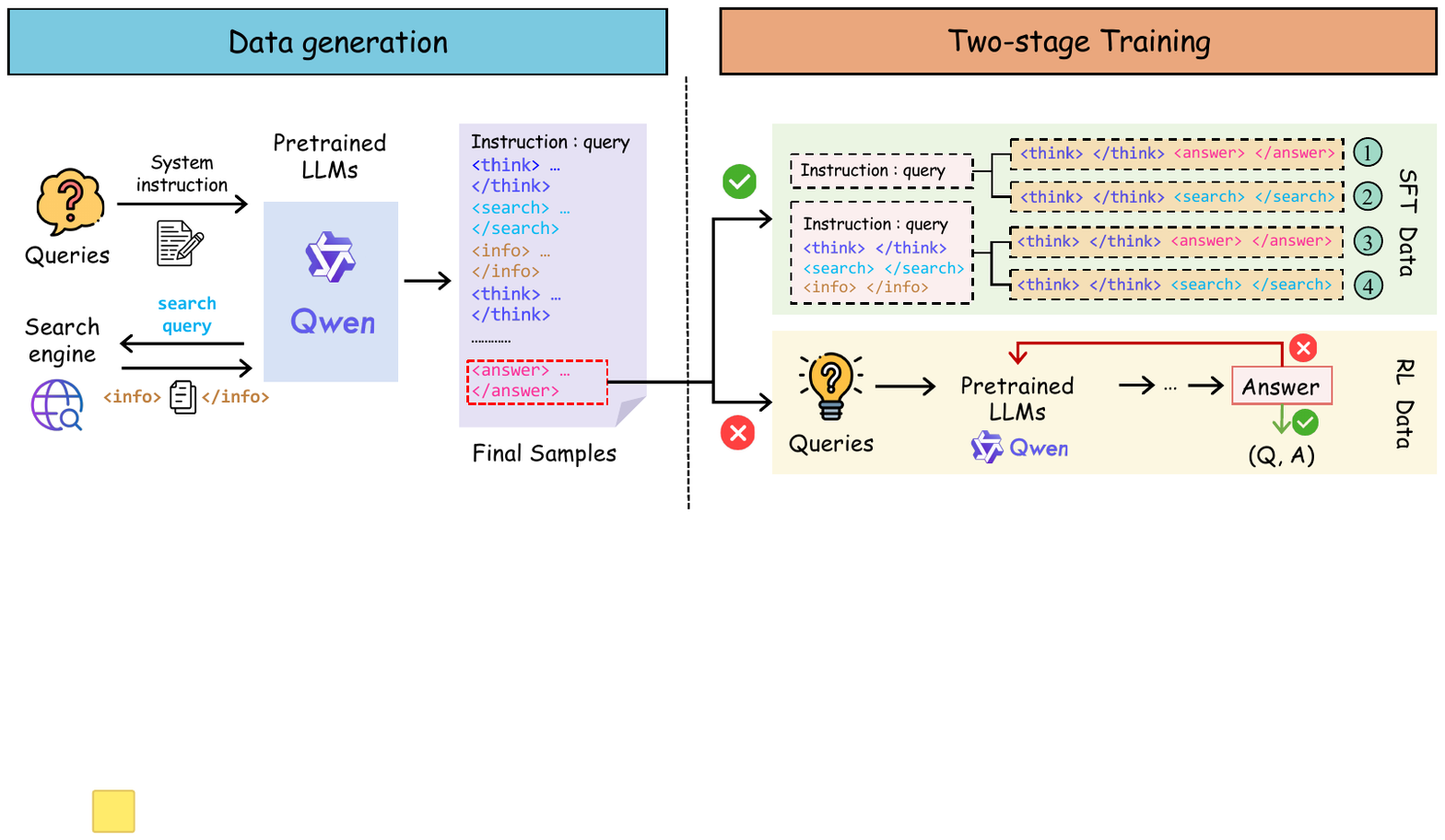}
  \caption{Overview of the RAG-R1 training framework, consisting of two stages: Format Learning Supervised Fine-Tuning (Section~\ref{sec:SFT}) and Retrieval-Augmented Reinforcement Learning (Section~\ref{sec:rl}), along with the data generation details.} 
  \label{Overall_framework}
\end{figure*}

In this section, we introduce the RAG-R1 training framework, which aims to empower LLMs with the ability to adaptively leverage internal and external knowledge during reasoning through two stages, i.e., Format Learning Supervised Fine-Tuning and Retrieval-Augmented Reinforcement Learning. 
Specifically, in the first stage, we thoughtfully generate samples that integrate reasoning and search, segmenting them into four categories.
We then apply SFT based on these segmented samples to equip LLMs with the ability to generate responses in a think-then-search format and leverage internal and external knowledge adaptively.
In the second stage, we conduct data selection and employ outcome-based RL with a retrieval environment to enhance the model's ability to reason and dynamically retrieve external knowledge to answer questions correctly. 
The overall framework is shown in Figure \ref{Overall_framework}.

\subsection{Format Learning Supervised Fine-Tuning}
\label{sec:SFT}
\subsubsection{SFT Data Generation}
\label{SFT_Data_Generation}
To equip LLMs with the ability to adaptively leverage internal and external knowledge during reasoning and respond in a think-then-search format, we first generate samples that integrate reasoning and search.
Specifically, the system instruction guides the LLM to perform reasoning between \think{$<$think$>$} and \think{$<$/think$>$} whenever it receives new information.
After completing the reasoning process, the LLM can generate search query between the designated search call tokens, \search{$<$search$>$} and \search{$<$/search$>$}, whenever external retrieval is necessary. 
The system will subsequently extract the search query and request an external search engine to retrieve relevant documents. 
The retrieved information is then appended to the existing sequence, enclosed within special retrieval tokens, \information{$<$information$>$} and \information{$<$/information$>$}, providing additional context for the next generation step. 
This process continues iteratively until reaching the maximum number of retrieval or the model generates a final answer enclosed within the special answer tokens, \answer{$<$answer$>$} and \answer{$<$/answer$>$}. 
The system instruction follows \citet{Search-R1} and is shown in Table \ref{instruction}.

\begin{table}[htbp]
\centering
\begin{tabular}{p{0.9\linewidth}}
\hline
Answer the given question. You must conduct reasoning inside \think{$<$think$>$} and \think{$<$/think$>$} first every time you get new information. After reasoning, if you find you lack some knowledge, you can call a search engine by \search{$<$search$>$} query \search{$<$/search$>$} and it will return the top searched results between \information{$<$information$>$} and \information{$<$/information$>$}. You can search as many times as your want. If you find no further external knowledge needed, you can directly provide the answer inside \answer{$<$answer$>$} and \answer{$<$/answer$>$}, without detailed illustrations. For example, \answer{$<$answer$>$} Beijing \answer{$<$/answer$>$}. Question: \question{question}\\
\hline
\end{tabular}
\caption{System instruction of SFT data generation.}
\label{instruction}
\end{table}

We use Qwen2.5-72B-Instruct as the generation model and employ a portion of the HotpotQA~\citep{yang2018hotpotqa} training dataset for generation tasks.

\subsubsection{SFT Data Segmentation and Training}
\label{SFT_Data_Segmentation}
After generating the samples, we selected those with correct answers and segment them into four categories according to specific segmentation points. 
As illustrated in Figure \ref{Overall_framework}, we segment the samples at corresponding points whenever the model needs to perform reasoning or retrieval, thereby preventing the model from generating retrieval documents. 
In this manner, a complete sample might be split into multiple smaller samples, which can then be classified into four distinct categories. More details can be found in Appendix \ref{SFT_Data_Segmentation_sample}.

By employing the first and second categories of samples, we train the model to respond by adaptively leveraging internal and external knowledge, respectively. 
In contrast, by utilizing the third and fourth categories of samples, we expect the model to perform reasoning and generate subsequent steps primarily based on external knowledge.
Notably, the output section of the samples is designed to exclude retrieved documents, which aids in preventing the model from generating hallucinations.

After segmentation, we apply SFT to these samples to train the model to generate responses in a think-then-search format and leverage internal and external knowledge adaptively.
This stage aims to develop a highly capable model that can respond in a think-then-search format, serving as a cold-start model to enhance the stability of the subsequent RL training stage.

\subsection{Retrieval-Augmented Reinforcement Learning}
\label{sec:rl}
\subsubsection{RL Data selection}
After Format Learning SFT, we obtain a model that can adaptively leverage internal and external knowledge during the reasoning process and respond in a think-then-search format.
To further enhance the model's reasoning ability and enable it to answer questions accurately, we begin with data selection to identify challenging yet answerable questions suitable for RL.

Specifically, we select the samples that generate incorrect answers in Section \ref{SFT_Data_Generation}. 
The questions of these samples present a relative challenge to the model. 
We subsequently filter out samples that are inherently unanswerable due to incomplete data retrieval or model limitations.
We implement stochastic sampling for Qwen2.5-72B-Instruct with the sampling temperature of 1.2 and the maximum number of retrieval to 10.
For each question, we conduct up to 10 rollouts and retain only samples containing at least one correct solution.
Finally, we obtain 2488 samples which are challenging yet answerable and randomly selected 25\% of the correctly answered samples in Section \ref{SFT_Data_Generation} to construct the final training dataset.

\subsubsection{RL with Retrieval}
We extend reinforcement learning to utilize an external retrieval system.
The RL objective function utilizing an external retrieval system $\mathcal{R}$ can be represented as follows: 
\begin{align*}
\underset{\pi_{\theta}} {max} &\mathbb{E}_{x \sim D, y \sim \pi_{\theta}(\cdot|x;\mathcal{R})} [r_{\phi}(x,y)] \\
&- \beta \mathbb{D}_{KL}[ \pi_{\theta}(y|x;\mathcal{R})|| \pi_{ref}(y|x;\mathcal{R}) ]
\end{align*}
where $\pi_{\theta}$ and $\pi_{ref}$ represent the policy model and the reference model, respectively, both of which are initialized from the SFT model, $r_{\phi}$ is the reward function and $\mathbb{D}_{KL}$ is KL-divergence. 
$x$ denote input samples drawn from the dataset $D$, and $y$ represent the generated outputs which is sampled from the policy model $\pi_{\theta}$ and retrieved from the retrieval system $\mathcal{R}$. 
The rollout process follows the same procedure detailed in Section \ref{SFT_Data_Generation}.
We employ the Proximal Policy Optimization (PPO)~\citep{PPO} algorithm to optimize the policy model, which is widely used in reinforcement learning due to its efficiency and reliability.

\textbf{Retrieval Masked Loss.} In our framework, the rollout sequence consists of both LLM-generated tokens and retrieved tokens from the external retrieval system. 
To prevent retrieved tokens from interfering with the inherent
reasoning and generation capabilities of the model, we implement a masked loss for these tokens.
By computing the policy gradient objective exclusively on the LLM-generated tokens and excluding the retrieved content from optimization, this approach stabilizes training while preserving the adaptability and benefits of retrieval-augmented generation.

\textbf{Reward Modeling.} The reward function serves as the primary training signal, which decides the optimization direction of RL. 
Inspired by \citet{deepseek-r1}, we adopt a rule-based reward system that includes final answer rewards to assess the accuracy of responses. 
For instance, in factual reasoning tasks, we adopt rule-based criteria such as exact string matching (EM) to evaluate correctness:
\begin{equation*}
r_{\phi}(x,y) = EM(a_{pre}, a_{gold})
\end{equation*}
where $a_{pre}$ represents the extracted final answer from response $y$ and $a_{gold}$ denotes the ground truth answer. This method focuses on ensuring that the final answer matches the expected truth precisely.
We avoid incorporating format rewards because the SFT model has already learned to respond in a think-then-search format. This strategy enables us to concentrate on the factual accuracy of the responses, leveraging the inherent structured approach of the model without imposing additional format-based conditions.
Furthermore, following \citet{deepseek-r1}, we do not apply neural reward models to avoid reward hacking and additional computational cost.

Utilizing the SFT model, which can adaptively leverage internal and external knowledge, as the cold-start model improves the stability of RL training.
Furthermore, by incorporating retrieval into outcome-based RL, we improve the model's reasoning capability and its ability to dynamically access external knowledge to accurately answer questions.

\section{Multi-query Parallelism}
Enabling the model to adaptively leverage internal and external knowledge during reasoning significantly extends its capability boundaries.
However, existing methods are constrained to a single-query approach, which presents two fundamental weaknesses: prohibitive latency from serial retrieval iterations, and inherent brittleness due to knowledge confinement. To overcome these limitations, we replace this fragile, sequential process with multi-query parallelism. This architectural shift is designed not merely to enhance capabilities, but to directly tackle the root causes of failure—it reduces latency by parallelizing retrieval and fortifies robustness by gathering diverse evidence simultaneously.

Specifically, we guide the model to generate multiple search queries whenever external retrieval is required. 
The external retrieval system then performs parallel searches with these queries and returns the results to the model in a JSON format, ensuring clear alignment between the search queries and the retrieved documents.
The adoption of multi-query parallelism directly confronts the two fundamental weaknesses of the single-query approach: prohibitive latency and inherent brittleness.
First, by executing retrievals in parallel, it dismantles the serial dependency that causes prohibitive latency, leading to a substantial reduction in overall inference time. Second, it shatters the knowledge confinement of a single query by gathering diverse evidence simultaneously. This enriched information base fortifies the model's reasoning process, dramatically enhancing its robustness against suboptimal queries and unrecoverable paths. This dual improvement is particularly salient for complex tasks like comparison-type questions, where synthesizing multiple perspectives is crucial for accuracy.

Table \ref{instruction_mq} illustrates the system instructions for generating multiple search queries.
We restrict the model to generate at most three parallel search queries simultaneously.
The retrieval system reorganized the search results in a JSON format which contains:
\begin{itemize}
\setlength{\itemsep}{0pt}
\setlength{\parskip}{0pt}
\setlength{\parsep}{0pt}
  \item \textbf{query}: A list of search queries generated by the model.
  \item \textbf{documents}: A list containing the at most three retrieved documents corresponding to the search queries.
\end{itemize}
Utilizing the above format allows the model to recognize the alignment between difference search queries and corresponding retrieved documents.
The remaining aspects of the training process are consistent with those outlined in Section \ref{training_framework}.
Detailed cases of multi-query methods are presented in the Appendix \ref{Case_Study}, showcasing its ability and efficiency to handle complex and multi-hop queries.

\begin{table}[htbp]
\centering
\begin{tabular}{p{0.9\linewidth}}
\hline
Answer the given question. You must conduct reasoning inside \think{$<$think$>$} and \think{$<$/think$>$} first every time you get new information. After reasoning, if you find you lack some knowledge, you can call a search engine by \search{$<$search$>$} query\_1,query\_2 \search{$<$/search$>$} and it will return the top searched results for each query between \information{$<$information$>$} and \information{$<$/information$>$}. You can search as many times as your want, using up to three queries each time. If you find no further external knowledge needed, you can directly provide the answer inside \answer{$<$answer$>$} and \answer{$<$/answer$>$}, without detailed illustrations. For example, \answer{$<$answer$>$} Beijing \answer{$<$/answer$>$}. Question: \question{question}\\
\hline
\end{tabular}
\caption{System instruction for generating multiple search queries.}
\label{instruction_mq}
\end{table}

\section{Experiments}
\subsection{Experimental Settings}

\textbf{Datasets and Evaluation Metrics.} 
We evaluate our models on seven benchmark datasets: 
(1) \textbf{General Question Answering}: NQ~\citep{Natural_Questions}, TriviaQA~\citep{JoshiTriviaQA2017}, and PopQA~\citep{mallen2023trustlanguagemodelsinvestigating}. 
(2) \textbf{Multi-Hop Question Answering}: HotpotQA~\citep{yang2018hotpotqa}, 2WikiMultiHopQA~\citep{xanh2020_2wikimultihop}, Musique~\citep{trivedi2021musique}, and Bamboogle~\citep{press-etal-2023-measuring}. 
HotpotQA is in-domain benchmark since a portion of its training sets is used for training while others serve as out-of-domain benchmarks to assess our model’s generalization capabilities.
For evaluation metric, we utilize Exact Match (EM) score following \citet{rankrag} and Retrieval Count (RC).

\textbf{Baselines.} 
We compare our method against the following baselines: 
(1) \textbf{Direct Inference}: Generate answers based on the model's internal knowledge.
(2) \textbf{Standard RAG}: Traditional retrieval-augmented generation systems. 
(3)\textbf{RAG-CoT Methods}: Integration of retrieval-augmented generation with chain-of-thought prompts, such as IRCoT~\citep{IRCoT} and Search-o1~\citep{Search-o1}. 
(4) \textbf{RAG-RL Methods}: Utilizing reinforcement learning to allow the model to autonomously perform retrieval during inference, such as Search-R1~\citep{Search-R1} and R1-Searcher~\citep{R1-Searcher}.

\textbf{Implementation Details.} 
We adopt Qwen-2.5-7B-Instruct as the base model for both our method and all baseline approaches, ensuring a fair comparison under the same architectural backbone.
For retrieval, we use the English Wikipedia provided by KILT~\citep{KILT} as retrieval corpus, segmented into 100-word passages with appended titles, totaling 29 million passages. 
We employ BGE-large-en-v1.5~\citep{bge_m3} as the text retriever and set the number of retrieved passages to 3 across all retrieval-based methods to ensure fair comparison. For evaluation, the number of retrieval iterations was left uncapped.

In Format Learning SFT, we collect 18994 samples for single-query training and 19303 samples for multi-query parallelism training. 
We perform full-parameter SFT for 5 epochs and chose the checkpoint with the lowest validation loss for evaluation and subsequent RL training.

In Retrieval-Augmented RL, we collect 5022 samples for single-query training and 6015 samples for multi-query parallelism training. We split 95\% of all the samples into a training set and used the remaining as a validation set. For the PPO variant, we set the learning rate of the policy LLM to 1e-6 and that of the value LLM to 1e-5. The training step is 500, with warm-up ratios of 0.285 and 0.015 for the policy and value models, respectively. We use Generalized Advantage Estimation (GAE)~\citep{GAE} with parameters $\lambda = 1$ and $\gamma = 1$. All the trainings are conducted on a single node with 8 A100 GPUs. 


\begin{table*}[t]
    \centering
    {
    \begin{tabular}{lcccccccc}
        \toprule
        \multirow{2}{*}{Methods} &   \multicolumn{3}{c}{\textbf{General QA}}   &   \multicolumn{4}{c}{\textbf{Multi-Hop QA}} &   \\
        \cmidrule(r){2-4}\cmidrule(r){5-8}
         &   NQ   &  PopQA  & TriviaQA  &  HotpotQA   &  2Wiki   &  Musique &  Bamboogle &   Avg.\\ 
        \midrule
        \midrule
        Direct Inference & 0.132  &  0.148 & 0.360  & 0.183  & 0.236  &  0.031 & 0.080 & 0.167 \\
        Standard RAG  & 0.328  &  0.353 &  0.476 &  0.284  & 0.253  &  0.048 & 0.152 & 0.271 \\
        IRCoT  & 0.183 & 0.328 & 0.434 & 0.276 & 0.356 & 0.060  & 0.144 & 0.254\\
        Search-o1  & 0.277 & 0.294 & 0.474 & 0.348 & 0.384  & 0.107 & 0.296 & 0.311\\
        Search-R1  & 0.387 & 0.422  & 0.531 & 0.377 & 0.351 & 0.135 & 0.376 & 0.368 \\
        R1-Searcher  & 0.404 & 0.410 & 0.522  & 0.442 & 0.513 & 0.158  & 0.368 & 0.402 \\
        \midrule
        RAG-R1-sq  & \textbf{0.429} & \underline{0.477} & \underline{0.599} &  \underline{0.492}  & \underline{0.520}  & \underline{0.187}  &  \textbf{0.440} & \underline{0.449} \\
        RAG-R1-mq  & \underline{0.423} & \textbf{0.479}  & \textbf{0.608} & \textbf{0.495}  & \textbf{0.563}  & \textbf{0.192} & \textbf{0.440} & \textbf{0.457}\\
        \bottomrule
    \end{tabular}
    \caption{Performance comparisons on QA benchmarks under the Exact Match metric. The best and second best results are \textbf{bold} and underlined, respectively. Our models, RAG-R1-sq and RAG-R1-mq, achieve substantial improvements over all baselines}
    \label{Main_results}
    }
\end{table*}

\subsection{Main results}
The main results comparing RAG-R1 with baseline methods across the seven datasets are presented in Table \ref{Main_results}. \emph{RAG-R1-sq} denotes our method operating in single-query mode while \emph{RAG-R1-mq} represents the overall framework incorporating multi-query parallelism.
From the results, we can obtain the following key observations:
\begin{itemize}
    \setlength{\itemsep}{0pt}
    \setlength{\parskip}{0pt}
    \setlength{\parsep}{0pt}
    \item \textbf{Significant performance improvements across all datasets.} Our method achieves significant improvements compared to all baseline methods across both General QA and mutil-hop QA benchmarks, including both CoT-based methods and RL-based methods. 
    Specifically, RAG-R1-mq outperforms the best RL-based method, R1-Searcher, by 13.7\% across all datasets.
    These results demonstrate that our approach enables the model to effectively utilize both internal and external knowledge throughout the reasoning process.
    
    \item \textbf{Effectiveness of Multi-query Parallelism.} RAG-R1-mq consistently outperforms all single-query RL-based methods, notably achieving a 1.8\% average gain over our own single-query counterpart, RAG-R1-sq.
    This result reveals a key insight: multi-query parallelism is not merely an efficiency optimization, but a direct remedy for the inherent brittleness of single-query methods. By shattering knowledge confinement, it empowers the model to synthesize diverse evidence, fundamentally bolstering its reasoning robustness against failures.
    Further details are available in Section \ref{Effects_of_Multiquery_Parallelism}.
    
    \item \textbf{Excellent generalization Ability.} Despite utilizing only a subset of the HotpotQA training data, our models achieve significant improvements on in-domain datasets and exhibit excellent generalization capabilities across out-of-domain datasets, such as 2WikiMultiHopQA and Musique. 
    This demonstrates that our models have effectively learned to reason and retrieve information for diverse questions, which proves the effectiveness of our approach across various scenarios requiring retrieval. Moreover, it can effectively extend to online search, as detailed in Section \ref{Online_Search}.
\end{itemize}

\begin{table*}[htbp]
    \centering
    \begin{tabular}{lccccc}
        \toprule
        Methods &  HotpotQA  &  2Wiki  &  Musique &  Bamboogle &   Avg.\\ 
        \midrule
        \midrule
        RAG-R1-sq & 0.492 & 0.520 & 0.187  & \textbf{0.440} & 0.410 \\
        RAG-R1-sq \textit{w/o SFT} & 0.415 & 0.406 & 0.138 & 0.312  & 0.318 \\
        RAG-R1-sq \textit{w/o RL} & 0.425 & 0.433  & 0.150 & 0.368  & 0.344 \\
        RAG-R1-sq \textit{w/o Filter} & 0.452 & 0.462 & 0.159 & 0.408 & 0.370 \\
        \midrule
        RAG-R1-mq  & \textbf{0.495} & \textbf{0.563}  & \textbf{0.192} & \textbf{0.440}  & \textbf{0.423} \\
        RAG-R1-mq  \textit{w/o SFT} & 0.413 & 0.423 & 0.129 & 0.392  & 0.339 \\
        RAG-R1-mq  \textit{w/o RL} & 0.427 & 0.490 & 0.143 & 0.424  & 0.371 \\
        RAG-R1-mq  \textit{w/o Filter} & 0.491 & 0.543 & 0.186 & 0.424 & 0.411 \\
        \bottomrule
    \end{tabular}
    \caption{Ablation study on SFT, RL and RL Data Selection. The RL-only method (w/o SFT) struggles to respond in a think-then-search format, while the SFT-only method (w/o RL) fails to enhance model performance through reasoning. Moreover, the performance decline observed in the w/o Filter method emphasizes the necessity of RL data selection.}
    \label{Ablation_study}
\end{table*}

\section{Further Analysis}
\subsection{Ablation Study}
To validate the effectiveness of our proposed training framework, we conduct a comprehensive ablation analysis of its key design elements.

\textbf{SFT and RL} As shown in Table \ref{Ablation_study}, \textit{w/o SFT} removes the initial format learning SFT while \textit{w/o RL} removes the entire RL training stage. 
The results demonstrate the necessity and effectiveness of both SFT and RL in our training framework, which together improve the model's performance. 
Specifically, \textit{w/o SFT} struggles to leverage internal and external knowledge and respond in a think-then-search format, resulting in decreased performance.
In contrast, \textit{w/o RL} restricts the model's capability to correctly answer questions through reasoning.
The considerable improvement achieved through RL highlight its ability to significantly enhance the model’s capability.

\textbf{RL Data Selection} We further validate the effectiveness of data selection during the RL process. 
We trained the models using all samples with incorrect answers and 25\% of the samples with correct answers in Section \ref{SFT_Data_Generation}, denoted as RAG-R1-sq \textit{w/o Filter} and RAG-R1-mq \textit{w/o Filter} respectively. 
The decline in performance indicates that careful data selection plays a crucial role in enhancing the effectiveness of RL training. 
Unanswerable samples do not facilitate the model's improvement and may even detrimentally impact the training process.

\begin{table}[htbp]
    \centering
    \small
    \begin{tabular}{lcccccc}
        \toprule
        Methods  &  \multicolumn{2}{c}{HotpotQA}   &   \multicolumn{2}{c}{2Wiki} &   \multicolumn{2}{c}{Avg.}  \\
        \cmidrule(r){2-3}\cmidrule(r){4-5}\cmidrule(r){6-7} & 
        Time & RI & Time & RI & Time & RI \\
        \midrule
        \midrule
        Search-R1  & 7.79 & 2.44  & 8.90 & 3.01 & 8.35 & 2.73 \\
        R1-Searcher & 10.98 & 2.31 & 10.93 & \textbf{2.40} & 10.96 & 2.36 \\
        \midrule
        RAG-R1-sq  & 7.69 & 2.14  & 8.72 & 2.72 & 8.21 & 2.43 \\
        RAG-R1-mq  & \textbf{6.72} & \textbf{1.89}  & \textbf{8.11} & 2.43 & \textbf{7.42} & \textbf{2.16} \\
        \bottomrule
    \end{tabular}
    \caption{Average time and retrieval iterations of different methods on two multi-hop datasets. RAG-R1-mq reduces the inference time by 9.6\% and the average retrieval iterations by 0.27 compared to RAG-R1-sq, demonstrating improved efficiency in multi-hop reasoning.}
    \label{time_and_retrieval_count}
\end{table}

\subsection{Effects of Multi-query Parallelism}
\label{Effects_of_Multiquery_Parallelism}
As shown in Table \ref{time_and_retrieval_count}, we record the average inference time (in seconds) and average retrieval iterations for different methods on HotpotQA and 2WikiMultiHopQA using A100 GPUs without employing any inference acceleration technique. 
The results show that RAG-R1-mq achieves the lowest inference time and requires the fewest retrieval iterations, indicating that multi-query parallelism can significantly enhance the efficiency of the model.
As shown in Table \ref{Main_results}, our multi-query approach (RAG-R1-mq) confirms its superior effectiveness. A detailed analysis against our single-query baseline (RAG-R1-sq) reveals this performance gain stems from two fundamental advantages:
(1) \textbf{More Efficient and Decisive Exploration}: The MQ model's parallel retrieval acts as a wide-ranging initial search, quickly identifying promising information pathways. This prevents the fruitless retrieval loops or premature terminations that plague the SQ model, which often lead to unrecoverable errors.
(2) \textbf{Adaptive Reasoning and Self-Correction}: When initial retrievals fail, the MQ model demonstrates a crucial capability for self-correction. It synthesizes the complete context from all parallel queries to intelligently formulate a more informed new query. In contrast, the SQ model, locked into its linear path, lacks this adaptive mechanism, rendering it brittle to an initially suboptimal query.
More case studies can be found in Appendix \ref{Case_Study}.
In essence, multi-query parallelism fosters a more robust and adaptive reasoning process, expanding the model's problem-solving capabilities beyond mere efficiency gains.

\begin{figure}[h] 
  \centering
  \includegraphics[width=1\linewidth]{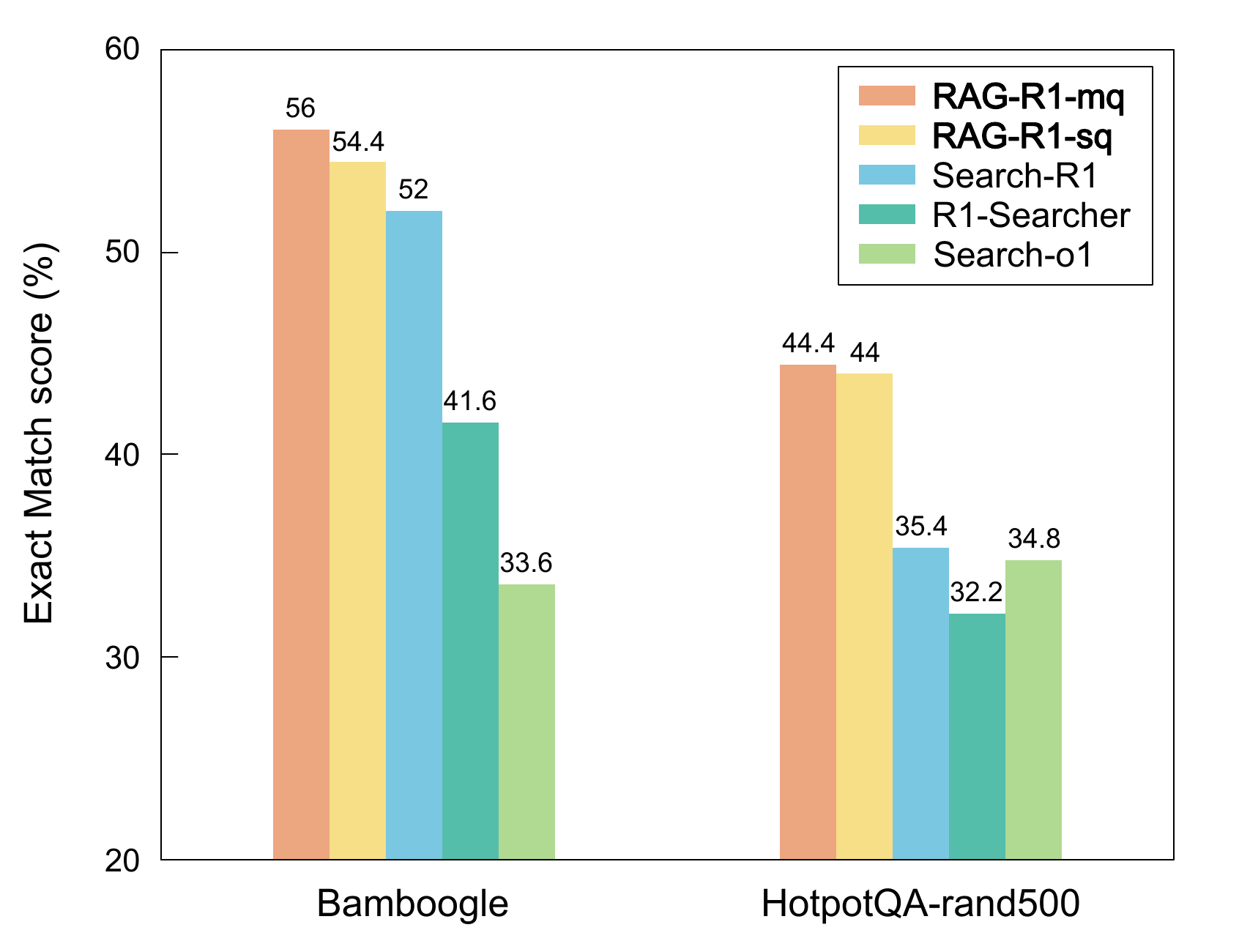}
  \caption{Performance comparison of RAG-R1 and three baselines within the online search scenario. Our models consistently deliver robust results across both offline and online settings, highlighting the strong generalization capabilities of our approach.} 
  \label{online_search_figure}
\end{figure}

\subsection{Generalization to Online Search}
\label{Online_Search}
Considering training efficiency and cost, we employ a local dense embedding retrieval system utilizing Wikipedia as a static external retrieval corpus throughout the training process.
This contrasts with most real-world applications, which depend on online web search. To demonstrate the generalization capability of RAG-R1 within this scenario, we assess our models' performance on two benchmarks: Bamboogle and a randomly selected set of 500 samples from HotpotQA, utilizing online web search—a setting not encountered during training.

Specifically, during inference, whenever retrieval is necessary, we leverage the Google API to perform real-time web searches and obtain relevant web pages. Subsequently, we utilize BeautifulSoup4 to scrape information from these pages and employ GPT-4o-mini to generate concise summaries, which are then incorporated into the reasoning process.
As illustrated in Figure \ref{online_search_figure}, RAG-R1-mq achieves the best EM score among all compared methods, highlighting its strong adaptability to online search scenario. These findings suggest that our approach equips the model to dynamically retrieve information during the reasoning process, rather than merely memorizing response formats.

\section{Conclusion} 
In this paper, we introduced RAG-R1, a novel training framework that enables LLMs to adaptively leverage internal and external knowledge, significantly enhancing their reasoning capabilities.
The cornerstone of our work is the integration of multi-query parallelism, an architectural innovation designed to directly address the prohibitive latency and inherent brittleness of conventional single-query methods, thereby bolstering reasoning robustness and reducing inference time.
Extensive experiments on seven QA benchmarks demonstrate the effectiveness of our method, which outperforms the strongest baseline by up to 13.7\% and decreases inference time by 11.1\%. This dual advancement confirms that our method achieves a superior trade-off, simultaneously boosting the model's reasoning robustness and inference efficiency.


\bibliography{aaai2026}

\appendix

\section{Implementation Details}
\label{Implementation_Details}
We adopt Qwen-2.5-7B-Instruct as the base model for both our method and all baseline approaches, ensuring a fair comparison under the same architectural backbone.
For retrieval, we use the English Wikipedia provided by KILT~\citep{KILT} as retrieval corpus, segmented into 100-word passages with appended titles, totaling 29 million passages. 
We employ BGE-large-en-v1.5~\citep{bge_m3} as the text retriever and set the number of retrieved passages to 3 across all retrieval-based methods to ensure fair comparison. 

In Format Learning SFT, we collect 18994 samples for single-query training and 19303 samples for multi-query parallelism training. 
We set the input length to 8192 and the output length to 2048.
We perform full-parameter SFT for 5 epochs and chose the checkpoint with the lowest validation loss for evaluation and subsequent RL training.

In Retrieval-Augmented RL, we collect 5022 samples for single-query training and 6015 samples for multi-query parallelism training. We split 95\% of all the samples into a training set and used the remaining as a validation set. For the PPO variant, we set the learning rate of the policy LLM to 1e-6 and that of the value LLM to 1e-5. The training step is 500, with warm-up ratios of 0.285 and 0.015 for the policy and value models, respectively. We use Generalized Advantage Estimation (GAE)~\citep{GAE} with parameters $\lambda = 1$ and $\gamma = 1$. All the trainings are conducted on a single node with 8 A100 GPUs. We configure the batch settings as follows: a total batch size of 512, a mini-batch size of 128, and a micro-batch size of 32.
The maximum sequence length is 8,192 tokens, with a maximum response length of 500 and retrieved content lengths set to a maximum of 1000 tokens for each search query.

For evaluation, the number of retrieval iterations was left uncapped, with a maximum response length of 1024 for each round and the temperature setting of 0.7.

\section{SFT Data Segmentation}
\label{SFT_Data_Segmentation_sample}
\onecolumn
\begin{table}[htbp]
\caption{Samples of SFT Data Segmentation}
\centering
\begin{tabular*}{\linewidth}{p{\linewidth}}
\toprule
\textbf{Raw Data}:\\
Question: Are the magazines, Nat Mags and Plaza Kvinna published in the same country? \\
\think{$<$think$>$}To determine if the magazines Nat Mags and Plaza Kvinna are published in the same country, I need to find out the countries of publication for both magazines.\think{$<$/think$>$}\\
\search{$<$search$>$} Nat Mags country of publication, Plaza Kvinna country of publication \search{$<$/search$>$}\\
\information{$<$information$>$}\\
\{\hspace{10pt}"query": $[$\\
\hspace{30pt}"Nat Mags country of publication", \\
\hspace{30pt}"Plaza Kvinna country of publication" $]$,\\
\hspace{15pt}"documents": $[$\\
\hspace{30pt}"Doc 1: Nat Mags   Nat Mags Nat Mags (short for National Magazine Company) was a British magazine publisher based in London...", \\
\hspace{30pt}"Doc 1: Plaza Kvinna   Plaza Kvinna Plaza Kvinna (meaning \"Plaza Woman\" in English) is a Swedish language monthly women's magazine published in Stockholm, Sweden..." \\
\hspace{15pt}$]$ \}\\
\information{$<$/information$>$}\\
\think{$<$think$>$}From the search results, Nat Mags is a British magazine publisher based in London, indicating that the magazines under Nat Mags are published in the United Kingdom. Plaza Kvinna, on the other hand, is a Swedish language monthly women's magazine published in Stockholm, Sweden. Therefore, these two magazines are not published in the same country.\think{$<$/think$>$}\\
\answer{$<$answer$>$} No \answer{$<$/answer$>$}\\
\midrule
\textbf{SFT Data after Segmentation}:\\
\textbf{Data 1}:\\
\textbf{input}: Question: Are the magazines, Nat Mags and Plaza Kvinna published in the same country?\\
\textbf{output}: \think{$<$think$>$}To determine if the magazines Nat Mags and Plaza Kvinna are published in the same country, I need to find out the countries of publication for both magazines.\think{$<$/think$>$}\\
\search{$<$search$>$} Nat Mags country of publication, Plaza Kvinna country of publication \search{$<$/search$>$}\\
\textbf{Data 2}:\\
\textbf{input}: Question: Are the magazines, Nat Mags and Plaza Kvinna published in the same country?\\
\think{$<$think$>$}To determine if the magazines Nat Mags and Plaza Kvinna are published in the same country, I need to find out the countries of publication for both magazines.\think{$<$/think$>$}\\
\search{$<$search$>$} Nat Mags country of publication, Plaza Kvinna country of publication \search{$<$/search$>$}\\
\information{$<$information$>$}\\
\{\hspace{10pt}"query": $[$\\
\hspace{30pt}"Nat Mags country of publication", \\
\hspace{30pt}"Plaza Kvinna country of publication" $]$,\\
\hspace{15pt}"documents": $[$\\
\hspace{30pt}"Doc 1: Nat Mags   Nat Mags Nat Mags (short for National Magazine Company) was a British magazine publisher based in London...", \\
\hspace{30pt}"Doc 1: Plaza Kvinna   Plaza Kvinna Plaza Kvinna (meaning \"Plaza Woman\" in English) is a Swedish language monthly women's magazine published in Stockholm, Sweden..." \\
\hspace{15pt}$]$ \}\\
\information{$<$/information$>$}\\
\textbf{output}: \think{$<$think$>$}From the search results, Nat Mags is a British magazine publisher based in London, indicating that the magazines under Nat Mags are published in the United Kingdom. Plaza Kvinna, on the other hand, is a Swedish language monthly women's magazine published in Stockholm, Sweden. Therefore, these two magazines are not published in the same country.\think{$<$/think$>$}\\
\answer{$<$answer$>$} No \answer{$<$/answer$>$}\\
\bottomrule
\end{tabular*}
\end{table}

\onecolumn
\section{Case Study}
\label{Case_Study}

\begin{table}[htbp]
\centering
\caption{Comparison of the format for single-query and multi-query parallelism}
\begin{tabular*}{\linewidth}{p{\linewidth}}
\toprule
\textbf{Question:} Which magazine was started first Arthur's Magazine or First for Women? \\
\midrule
\textbf{Ground Truth:} Arthur's Magazine \\
\midrule
\textbf{Single-query:} \\
\think{$<$think$>$} To determine which magazine was started first between Arthur's Magazine and First for Women, I need to find the founding years of both magazines. \think{$<$/think$>$}\\
\search{$<$search$>$} founding year of Arthur's Magazine \search{$<$/search$>$}\\
\information{$<$information$>$}\\Doc 1: Arthur (magazine)   moved Arthur's headquarters from Los Angeles to New York, the seat of North America's publishing industry. On March 6, 2011, Jay Babcock announced that the magazine would cease to exist in any form as of March 15, 2011, though its archive and store would remain active for an unspecified period thereafter...\\
\information{$<$/information$>$}\\
\hdashrule[0.5ex]{\linewidth}{0.5pt}{5pt} \\
\think{$<$think$>$} From the search results, I found that Arthur's Magazine was published from 1844 to 1846. Now, I need to find the founding year of First for Women. \think{$<$/think$>$}\\
\search{$<$search$>$} founding year of First for Women magazine \search{$<$/search$>$}\\
\information{$<$information$>$}\\
Doc 1: First for Women - Wikipedia   First for Women is an American women's interest magazine published by Bauer Media Group. It was founded in 1978 and is based in New York City. The magazine focuses on topics such as health, beauty, relationships, and personal finance. It is known for its practical advice and real-life stories. First for Women has a circulation of over 1 million readers and is available in both print and ...\\
\information{$<$/information$>$}\\
\hdashrule[0.5ex]{\linewidth}{0.5pt}{5pt} \\
\think{$<$think$>$} First for Women was founded in 1978. Comparing the founding years, Arthur's Magazine (1844) is clearly older than First for Women (1978). \think{$<$/think$>$}\\
\answer{$<$answer$>$} Arthur's Magazine \answer{$<$/answer$>$}\\
\midrule
\textbf{Multiple-query parallelism:} \\
\think{$<$think$>$} To determine which magazine was started first, I need to compare the founding years of both "Arthur's Magazine" and "First for Women." \think{$<$/think$>$}\\
\search{$<$search$>$} \textbf{Arthur's Magazine founding year, First for Women founding year} \search{$<$/search$>$} \\
\information{$<$information$>$}\\
\{\hspace{10pt}"query": $[$\\
\hspace{30pt}"Arthur's Magazine founding year", \\
\hspace{30pt}"First for Women founding year" $]$,\\
\hspace{15pt}"documents": $[$\\
\hspace{30pt}"Doc 1: Arthur's Magazine   Arthur's Magazine Arthur's Magazine (1844-1846) was an American literary periodical published in Philadelphia in the 19th century. Edited by Ti...", \\
\hspace{30pt}"Doc 1: First woman   firsts Doc 2: First for Women   First for Women First for Women is a woman's magazine published by Bauer Media Group in the USA. The magazine was started in 1989. It is based in Englewood Cliffs, New Jersey..." \\
\hspace{15pt}$]$ \}\\
\information{$<$/information$>$}\\
\hdashrule[0.5ex]{\linewidth}{0.5pt}{5pt} \\
\think{$<$think$>$} From the search results: - "Arthur's Magazine" was published from 1844 to 1846. - "First for Women" was started in 1989. Therefore, "Arthur's Magazine" was started first. \think{$<$/think$>$}\\
\answer{$<$answer$>$} Arthur's Magazine \answer{$<$/answer$>$}\\
\bottomrule
\end{tabular*}
\end{table}

\onecolumn
\begin{longtable}{p\linewidth}
\caption{Comparison of the effectiveness between single-query and multi-query parallelism} \\
\midrule
\textbf{Question:} Are both Dictyosperma, and Huernia described as a genus? \\
\midrule
\textbf{Ground Truth:} Yes \\
\midrule
\textbf{Single-query:} \\
\think{$<$think$>$} To determine if both Dictyosperma and Huernia are described as genera, I need to verify the taxonomic classification of each. \think{$<$/think$>$}\\
\search{$<$search$>$} \textcolor{red}{\textbf{Dictyosperma genus}} \search{$<$/search$>$}\\
\information{$<$information$>$}\\Doc 1: Dictyospermum   Dictyospermum Dictyospermum is a genus of monocotyledonous flowering plants in the dayflower family, first described in 1849. It is native to Africa ...\\
\information{$<$/information$>$}\\
\hdashrule[0.5ex]{\linewidth}{0.5pt}{5pt} \\
\think{$<$think$>$} The search results indicate that Dictyosperma is not clearly identified as a genus in the provided information. It seems to be referred to as "Polyspatha" in the context of species. I will need to conduct another search specifically for "Dictyosperma" to confirm its status as a genus. \think{$<$/think$>$}\\
\search{$<$search$>$} Huernia genus \search{$<$/search$>$}\\
\information{$<$information$>$}\\
Doc 1: Huernia have shown the genus to be monophyletic, and most closely related to the genus "Tavaresia", and to a widespread branch of stapeliads comprising the genera "Orbea", "Piaranthus" and "Stapelia" ...\\
\information{$<$/information$>$}\\
\hdashrule[0.5ex]{\linewidth}{0.5pt}{5pt} \\
\think{$<$think$>$}The search results confirm that Huernia is indeed described as a genus, with its first description dating back to 1810. Now, I need to confirm the status of Dictyosperma. \think{$<$/think$>$}\\
\search{$<$search$>$} \textcolor{red}{\textbf{Dictyosperma genus}} \search{$<$/search$>$}\\
\information{$<$information$>$}\\Doc 1: Dictyospermum   Dictyospermum Dictyospermum is a genus of monocotyledonous flowering plants in the dayflower family, first described in 1849. It is native to Africa ...\\
\information{$<$/information$>$}\\
\hdashrule[0.5ex]{\linewidth}{0.5pt}{5pt} \\
\think{$<$think$>$} The search results do not provide clear evidence that Dictyosperma is a genus. Instead, it seems to be associated with the genus Polyspatha. Given the information available, I cannot confirm that both Dictyosperma and Huernia are described as genera. \think{$<$/think$>$}\\
\answer{$<$answer$>$} \textcolor{red}{\textbf{No}} \answer{$<$/answer$>$}\\
\midrule
\textbf{Multiple-query parallelism:} \\
\think{$<$think$>$} To determine if both Dictyosperma and Huernia are described as genera, I need to verify their taxonomic status. \think{$<$/think$>$}\\
\search{$<$search$>$} \textcolor{red}{\textbf{Dictyosperma genus, Huernia genus}} \search{$<$/search$>$} \\
\information{$<$information$>$}\\
\{\hspace{10pt}"query": $[$\\
\hspace{30pt}"Dictyosperma genus", \\
\hspace{30pt}"Huernia genus" $]$,\\
\hspace{15pt}"documents": $[$\\
\hspace{30pt}"Doc 1: Dictyospermum   Dictyospermum Dictyospermum is a genus of monocotyledonous flowering plants in the dayflower family, first described in 1849. It is native to Africa ...", \\
\hspace{30pt}"Doc 1: Huernia have shown the genus to be monophyletic, and most closely related to the genus "Tavaresia", and to a widespread branch of stapeliads comprising the genera "Orbea", "Piaranthus" and "Stapelia" ..." \\
\hspace{15pt}$]$ \}\\
\information{$<$/information$>$}\\
\hdashrule[0.5ex]{\linewidth}{0.5pt}{5pt} \\
\think{$<$think$>$} From the search results, it appears that:\\
- Dictyosperma is not explicitly mentioned as a genus, but there is a genus named Dictyospermum, which might be a similar spelling.\\
- Huernia is indeed described as a genus within the family Apocynaceae, subfamily Asclepiadoideae.\\
I should confirm the status of Dictyosperma. \think{$<$/think$>$}\\
\search{$<$search$>$} \textcolor{red}{\textbf{Dictyosperma plant}} \search{$<$/search$>$} \\
\information{$<$information$>$}\\
\{\hspace{10pt}"query": $[$\\
\hspace{30pt}"Dictyosperma plant", \\
\hspace{30pt} $]$,\\
\hspace{15pt}"documents": $[$\\
\hspace{30pt}"Doc 1: Dictyoloma   Dictyoloma Dictyoloma is a genus of flowering plants that belongs to the family Rutaceae ...", \\
\hspace{15pt}$]$ \}\\
\information{$<$/information$>$}\\
\hdashrule[0.5ex]{\linewidth}{0.5pt}{5pt} \\
\think{$<$think$>$} Based on the additional search, Dictyosperma is confirmed to be a monotypic genus of flowering plant in the palm family, found in the Mascarene Islands. \\
Both Dictyosperma and Huernia are described as genera. \think{$<$/think$>$}\\
\answer{$<$answer$>$} \textcolor{red}{\textbf{Yes}} \answer{$<$/answer$>$}\\
\bottomrule
\end{longtable}
\end{document}